\documentclass[conference,hidelinks]{IEEEtran}
\usepackage[font=small]{caption} 
\usepackage{graphicx,color}
\usepackage{color}
\usepackage{xcolor}
\usepackage{setspace} 
\usepackage{amsmath} 
\usepackage{amsmath,amsthm,amssymb}
\usepackage{multirow}
\usepackage{mathtools}
\usepackage{hhline}
\usepackage{verbatim}
\usepackage{algorithm}
\usepackage{algcompatible}
\usepackage{multirow}
\usepackage{algorithmicx}
\usepackage{algpseudocode}
\usepackage{algpseudocode}
\usepackage{caption}[=v3.5]
\usepackage{subcaption}
\usepackage{cases}

\newcommand{\qty}[1]{#1}

\usepackage{hyperref}
\usepackage[]{glossaries-extra}

\usepackage{tabularx}
\newcolumntype{Y}{>{\centering\arraybackslash}X}

\usepackage{orcidlink}

\usepackage{soul}

\setabbreviationstyle[acronym]{long-short}

\newacronym{KTH}{KTH}{KTH Royal Institute of Technology}
\newacronym{RISE}{RISE}{Research Institute of Sweden}
\newacronym[shortplural={UAVs}, firstplural={Unmanned Aerial Vehicles (UAVs)}]{UAV}{UAV}{Unmanned Aerial Vehicle}
\newacronym{MAV}{MAV}{Micro Aerial Vehicle}
\newacronym{UAS}{UAS}{Unmanned Aerial System}
\newacronym{RPAV}{RPAV}{Remotely-Piloted Aerial Vehicle}
\newacronym{VTOL}{VTOL}{Vertical Take-Off and Landing}
\newacronym{CNN}{CNN}{Convolutional Neural Network}
\newacronym{AI}{AI}{Artificial Intelligence}
\newacronym{DL}{DL}{Deep Learning}
\newacronym{WiFi}{Wi‑Fi}{Wireless Fidelity}
\newacronym{SAR}{SAR}{Search And Rescue}
\newacronym{IoT}{IoT}{Internet of Things}
\newacronym{PID}{PID}{Proportional Integral Derivative}
\newacronym{SC}{SC}{Split Computing}
\newacronym{HW}{HW}{Hardware}
\newacronym{SW}{SW}{Software}
\newacronym{GNSS}{GNSS}{Global Navigation Satellite System}
\newacronym{SLAM}{SLAM}{Simultaneous Localization and Mapping}
\newacronym{MPC}{MPC}{Model Predictive Control}
\newacronym{EKF}{EKF}{Extended Kalman Filter}

\newacronym{PRM}{PRM}{Probabilistic Roadmaps}
\newacronym{RRT}{RRT}{Rapidly Exploring Random Trees}
\newacronym{PSO}{PSO}{Particle Swarm Optimisation}
\newacronym{APF}{APF}{Artificial Potential Field}
\newacronym{VFF}{VFF}{Virtual Force Field}
\newacronym{VFH}{VFH}{Vector Field Histogram}
\newacronym{ND}{ND}{Nearness Diagram}
\newacronym{CG}{CG}{Closest-Gap}
\newacronym{EG}{EG}{Escape Gap}
\newacronym{SIFT}{SIFT}{Scale-Invariant Feature Transform}
\newacronym{SURF}{SURF}{Speed Up Robust Features}
\newacronym{OF}{OF}{Optical Flow}
\newacronym{IMU}{IMU}{Inertial Measurement Unit}
\newacronym{CBF}{CBF}{Control Barrier Function}

\newacronym{API}{API}{Application Programming Interface}
\newacronym{RTOS}{RTOS}{Real-Time Operating System}
\newacronym[shortplural={PCBs}, firstplural={Printed Circuit Boards (PCBs)}]{PCB}{PCB}{Printed Circuit Board}
\newacronym{COTS}{COTS}{Commercial Off-the-Shelf}
\newacronym[shortplural={MCUs}, firstplural={Microcontroller Units}]{MCU}{MCU}{Microcontroller Unit}
\newacronym{OOT}{OOT}{Out-of-Tree}
\newacronym{PULP}{PULP}{Parallel Ultra Low Power}
\newacronym{ULP}{ULP}{Ultra Low Power}
\newacronym{SWaP}{SWaP}{Size, Weight and Power}
\newacronym{FOV}{FOV}{Field of View}
\newacronym{ToF}{ToF}{Time-of-Flight}
\newacronym{PWM}{PWM}{Pulse Width Modulation}
\newacronym{POST}{POST}{Power-On-Self-Test}
\newacronym{SoC}{SoC}{System on Chip}
\newacronym{CRTP}{CRTP}{Crazyflie Transfer Protocol}
\newacronym{CPX}{CPX}{Crazyflye Package eXchange}
\newacronym{FC}{FC}{Fabric Controller}
\newacronym{CL}{CL}{Cluster}
\newacronym{DMA}{DMA}{Direct Memory Access}
\newacronym{MTU}{MTU}{Maximum Transmission Unit}

\newacronym{mAP}{mAP}{mean Average Precision}
\newacronym{AP}{AP}{Average Precision}
\newacronym{PR}{PR}{Precision-Recall}
\newacronym[shortplural={BBs}, firstplural={Bounding Boxes}]{BB}{BB}{Bounding Box}
\newacronym{IoU}{IoU}{Intersection over Union}
\newacronym{TP}{TP}{True Positive}
\newacronym{FP}{FP}{False Positive}
\newacronym{TN}{TN}{True Negative}
\newacronym{FN}{FN}{False Negative}

\newacronym[shortplural={SMs}, firstplural={State Machines}]{SM}{SM}{State Machine}
\newacronym{SSD}{SSD}{Single-Shot Detector}

\usepackage{enumitem}

\begin{document}
\bstctlcite{IEEEexample:BSTcontrol}
\title{AI and Vision based Autonomous Navigation of Nano-Drones in Partially-Known Environments \vspace*{-0.05in}}

\author{
\IEEEauthorblockN{Mattia Sartori\IEEEauthorrefmark{1}\orcidlink{0009-0007-1187-3543}, Chetna Singhal\IEEEauthorrefmark{2}\orcidlink{0000-0002-4712-8162}, Neelabhro Roy\IEEEauthorrefmark{1}\orcidlink{0000-0002-5777-7780}, Davide Brunelli\IEEEauthorrefmark{3}\orcidlink{0000-0001-5110-6823}, James Gross\IEEEauthorrefmark{1}\orcidlink{0000-0001-6682-6559}\\
 			\IEEEauthorrefmark{1}{KTH Royal Institute of Technology, Sweden.} \,\,\, 
    \IEEEauthorrefmark{2}{Inria France.} \,\,\, 
    		\IEEEauthorrefmark{3}{University of Trento, Italy.}\\
 		}\vspace*{-0.25in}}

\maketitle
\begin{abstract}
The miniaturisation of sensors and processors, the advancements in connected edge intelligence, and the exponential interest in Artificial Intelligence are boosting the affirmation of autonomous nano-size drones in the Internet of Robotic Things ecosystem. However, achieving safe autonomous navigation and high-level tasks such as exploration and surveillance with these tiny platforms is extremely challenging due to their limited resources.
This work focuses on enabling the safe and autonomous flight of a pocket-size, 30-gram platform called Crazyflie 2.1 in a partially known environment. We propose a novel AI-aided, vision-based reactive planning method for obstacle avoidance under the ambit of Integrated Sensing, Computing and Communication paradigm. We deal with the constraints of the nano-drone by splitting the navigation task into two parts: a deep learning-based object detector runs on the edge (external hardware) while the planning algorithm is executed onboard. 
The results show the ability to command the drone at $\sim8$ frames-per-second and a model performance reaching a COCO mean-average-precision of $60.8$. Field experiments demonstrate the feasibility of the solution with the drone flying at a top speed of \qty{1}{m/s} while steering away from an obstacle placed in an unknown position and reaching the target destination. The outcome highlights the compatibility of the communication delay and the model performance with the requirements of the real-time navigation task. 
We provide a feasible alternative to a fully onboard implementation that can be extended to autonomous exploration with nano-drones.
\end{abstract}

{\IEEEkeywords Obstacle Avoidance, Reactive Planning, Vision-based Autonomous Navigation, Nano Drones.}

\section{Introduction}
\label{sec:introduction}
The Internet of Robotic Things (IoRT) is an emerging Internet of Things paradigm where robots are provided advanced situational awareness thanks to sensors and data analytics methods implemented onboard and on the edge \cite{iort}.
Among other platforms, small-sized drones have attracted increasing interest due to their high flexibility and cost-effectiveness. Use-case specific payload makes these platforms viable in tasks such as exploration, search and rescue, infrastructure inspection, precision agriculture, and many more \cite{drones_in_iot}.

The ongoing endeavour is to empower Unmanned Aerial Vehicles (UAVs) with advanced autonomy while tackling their intrinsic limitations using the Integrated Sensing, Computing and Communication (ISCC) technology \cite{dcoss_2023}. A fundamental challenge for mobile robots and UAVs is to perform safe autonomous navigation between two points in the presence of obstacles or under variable weather conditions~\cite{albanese_drone_rain}. This is even more challenging for the tiny UAVs considering Size, Weight and Power constraints~\cite{dcoss_2020}.

We have developed an ISCC framework for the safe navigation of a nano-drone, called Crazyflie, relying only on visual input and an Inertial Measurement Unit. Vision comes from a monocular, \gls{ULP}, low-resolution, greyscale camera with limited \gls{FOV}. We circumvent the lack of depth information (knowledge of the relative distance from potential obstacles) by empowering the drone with a reactive planning method for obstacle avoidance based on visual obstacle detection. Moreover, we explore the performance of splitting the navigation task between onboard resources and external \gls{HW} capable of more intensive computing. This strategy resembles the concept of Split Computing, which can be defined as the distribution of the computing load across a mobile device (e.g. a UAV) and an edge server~\cite{singhal_infocom2024}. Usually, the split is employed in the execution of a Deep Neural Network (DNN) that is divided into sub-models that run separately on two or more devices~\cite{singhal_secon2024}  \cite{a3_review}. In our work, we offload the AI inference task that processes the available visual sensory input, and we use the resulting predictions on the nano-drone to compute iteratively the desired actions needed to reach a waypoint while avoiding collisions.

We have implemented the framework to execute the high-level tasks for low-power UAVs by leveraging  edge device capabilities. 
We summarise the key contributions as follows:
\begin{itemize}[itemsep=0mm,leftmargin=2mm]  
    \item Developed a minimal working solution for nano-drones (Crazyflie 2.1) that can be adapted for autonomous surveillance and exploration of a semi-unknown environment.
    \item Created a low-resolution, greyscale image dataset for supervised learning and fine-tuning of an object detection model.
    \item Proposed a heuristic, vision-based, reactive planning method for obstacle avoidance that does not rely on distance information. We have validated it with field experiments.
    \item Analysed the effects of splitting the navigation computational task between the edge and nano-drone platform.
\end{itemize}

\section{Related Work}
\label{sec:rel-work}
Drones are an integral part of IoRT and can be classified according to size, weight, and power, as in \autoref{tab:drone_size}~\cite{palossi_open_2019}. In our work, we employ a nano drone called \emph{Crazyflie 2.1} mounting an expansion board called \emph{AI-deck} based on a GAP8 processor~\cite{palossi_64-mw_2019}.

\begin{table}[!ht]
  \begin{center}
  {\small
      \vspace{-3mm}
    \caption{Rotorcraft UAVs taxonomy by vehicle class-size}
    \vspace{-3mm}
    \label{tab:drone_size}
    \begin{tabularx}{\linewidth}{YYYY}
        \hline
        \textbf{Vehicle Class} & $\oslash$\textbf{:Weight [\emph{cm:Kg}]} & \textbf{Power [\emph{W}]} & \textbf{On-board Device}\\
        \hline
        \emph{std-size}   & $\sim50$:$\geq1$      & $\geq100$   & Desktop \\
        \emph{micro-size} & $\sim25$:$\sim0.5$  & $\sim50$  & Embedded \\
        \textbf{\emph{nano-size}}  & $\sim10$:$\sim0.05$ & $\sim5$   & MCU \\
        \emph{pico-size}  & $\sim2$:$\sim0.005$ & $\sim0.1$ & ULP \\
        \hline
    \end{tabularx}
    \vspace{-3mm}
  }
  \end{center}
\end{table}

The ISCC use cases, such as autonomous navigation and driving, depend on sensing environmental information using an array of onboard sensors (e.g. camera, LiDAR, and Radar)~\cite{dcoss_2020}. Previously, range measurements and wall-following behaviour were used for navigation and collision avoidance with Crazyflie~\cite{mcguire_minimal_2019,liu_adaptive_2023}. 
The AI-deck on our drone instead mounts a \gls{ULP} camera, which does not provide depth information. Visual sensors are compact, light, and cheap and can provide rich environmental information which, however, requires real-time processing to allow safe navigation. This is challenging to achieve onboard and can leverage computing resources in an edge-offloading architecture~\cite{singhal_infocom2024}.

Visual navigation methods with Crazyflie included optical flow estimation and balancing \cite{bouwmeester_nanoflownet_2023} and depth estimation \cite{end_to_end_zhang}. However, the approach in \cite{end_to_end_zhang} is computationally expensive and can run at only \qty{1.24}{FPS} at full resolution, thus not suitable for real-time flight control.
SoA work also focused on embedding tiny DNNs trained end-to-end on the GAP8 aboard the drone to achieve different tasks such as pose estimation ~\cite{palossi_fully_2022}, object detection ~\cite{lamberti_bio-inspired_2023} and navigation ~\cite{a1_review}.
The authors of ~\cite{lamberti_bio-inspired_2023} perform obstacle detection with a DNN based on SSD-Mobilenet V2 ~\cite{sandler_mobilenetv2_2018} achieving a throughput of \qty{4.3}{FPS} with their most quantised network. They do not use the detection information for obstacle avoidance. The methods in ~\cite{a1_review} and ~\cite{pulp_dronet_v3} extend the PULP-dronet framework ~\cite{palossi_64-mw_2019} allowing safe autonomous navigation in an unknown environment. 
They implement a reactive control approach that steers the drone away from the direction of the highest estimated collision probability without considering a target location to reach. 
In our work instead, we want to exploit the information on the detected obstacle for reactive planning of the drone's motion around the obstacle and towards defined waypoints.
Moreover, the previously cited tasks are computationally expensive relative to the onboard HW and leave very few resources available to analyse additional sensor inputs, execute additional navigation/control tasks and run other AI models ~\cite{pulp_dronet_v3}. 
Even if there are known limitations in the communications between an edge platform and a nano drone ~\cite{a2_review}, split computing and edge offloading can significantly boost the drone’s perception capabilities ~\cite{a3_review} and contribute to spare onboard resources.

\newcolumntype{C}{ >{\centering\arraybackslash} m{0.185in} }
\newcommand{\tick}{\textbf{\textcolor{green}{\checkmark}}}
\newcommand{\cross}{\textbf{\textcolor{red}{$\times$}}}

\begin{table}[!htb]
    {\small
    \renewcommand{\tabcolsep}{0.5pt}
    \footnotesize
    \begin{center}
    \caption{Feature comparison with state-of-the-art.}
    \vspace{-3mm}
    \label{tab:comparison}
    \begin{tabularx}{\linewidth}{| >{\raggedright\arraybackslash}X |
                            >{\centering\arraybackslash} m{0.21in}
                            |*{8}{C|} 
                            >{\centering\arraybackslash} m{0.21in}|}
        \hline
        \rule{0pt}{1.5ex}Features&\!\!\cite{a3_review}&\!\! \cite{mcguire_minimal_2019}&\!\!\cite{lamberti_bio-inspired_2023}&\!\! \cite{soria_predictive_2021}&\!\!\cite{liu_adaptive_2023}&\!\!\cite{a1_review}&\!\!\cite{pulp_dronet_v3}&\!\!\cite{bouwmeester_nanoflownet_2023}&\!\!\cite{palossi_64-mw_2019}&\!\textbf{Our}\!\!\\
        \hline
        \rule{0pt}{1.5ex}Vision based navigation & \tick & \cross & \cross & \tick & \cross & \tick & \tick & \tick & \tick & \tick \\
        \hline
        \rule{0pt}{1.5ex}AI-based object detection & \cross & \cross & \tick & \cross & \cross & \cross & \cross & \cross & \cross & \tick \\
        \hline
        \rule{0pt}{1.5ex}Computation (On/Off-board) &Split&On&On&Off&Off&On&On&On&On&Split\\
        \hline
        \rule{0pt}{1.5ex}Onboard state estimation & \tick & \tick & \tick & \cross & \cross & \tick & \tick & \tick & \tick & \tick \\
        \hline
        \rule{0pt}{1.5ex}Waypoint navigation & \cross & \cross & \cross & \tick & \tick & \cross & \cross & \cross & \cross & \tick \\
        \hline
        \rule{0pt}{1.5ex}AI-based obstacle avoidance & \cross & \cross & \cross & \cross & \tick & \tick & \tick & \tick & \tick & \tick \\
        \hline
        \rule{0pt}{1.5ex}Unknown obstacle position & \cross & \tick & \cross & \cross & \tick & \tick & \tick & \tick & \tick & \tick \\
        \hline
        \rule{0pt}{1.5ex}Latency analysis & \tick & \cross & \cross & \cross & \cross & \cross & \cross & \cross & \cross & \tick \\
        \hline
        \rule{0pt}{1.5ex}Planning and control focus & \cross & \cross & \cross & \tick & \tick & \cross & \cross & \cross & \cross & \tick \\
        \hline
    \end{tabularx}
    \vspace{-4mm}
    \end{center}
    }
\end{table}

In \autoref{tab:comparison}, we provide a feature comparison and fundamental differences in the objectives of our proposed framework compared to the state-of-the-art.
We provide a feasible alternative to an onboard solution by exploiting the support of an edge device to run an obstacle-detection task based on SSD-Mobilenet. We use the information on the detected obstacles in the planning algorithm for safe navigation. Additionally, we focus on the planning and control of the drone combining obstacle avoidance and the navigation towards a waypoint.

\section{Autonomous Nano-drone Navigation Solution}
\label{sec:autonomous_navigation}
\begin{figure}[tb]
  \begin{center}
    \includegraphics[width=\linewidth]{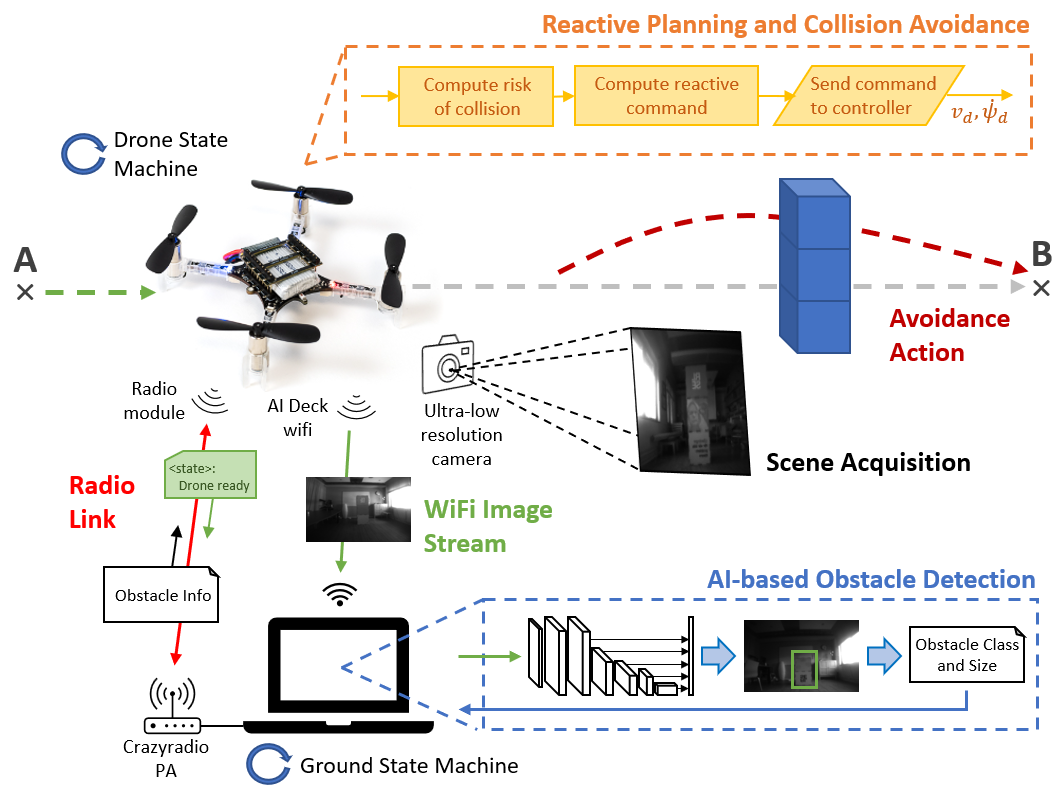}
  \end{center}
    \vspace{-3mm}
  \caption{\textbf{High-level pipeline representation.} Nano-drone flies from point A to B and avoids obstacles using an AI-aided, vision-based reactive planning. The navigation task is split: obstacle detection on the external HW, planning on the UAV.}
    \vspace{-5mm}
  \label{fig:background_pipeline}
\end{figure}

We use the Crazyflie 2.1 quadcopter from Bitcraze AB (\qty{30}{-gram}, $10\times10$ cm dimension, $\leq$ \qty{8}{W} power envelope, and $\sim$ 7 minutes flight time \cite{palossi_64-mw_2019}) with two \glspl{MCU}, STM32 (state estimation and control tasks) and nRF51822 (radio and power management). 
We extend the quadcopter with two decks, the Flow-deck (relative motion detection) and the 
AI-deck (GAP8 high-level computing visual engine, greyscale QVGA camera, and an ESP32-based Wi-Fi transceiver). 
A long-range open USB radio dongle called Crazyradio is used for wireless communication between the drone and external hardware.

Our framework overcomes the memory and computational resource limitations while providing miniaturised and energy-efficient intelligence and autonomy.
The key elements of the autonomous navigation solution are (\autoref{fig:background_pipeline}) 1) synchronisation and state information exchange; 2) scene acquisition and transmission; 3) deep-learning based object detection; and 4) reactive planning and collision avoidance.

\subsection{State-information and synchronisation}
We have implemented a \gls{SM} on the drone and the Offloading Computing Unit (OCU) to synchronise the execution of the split navigation task and to exchange state information over a radio link. 
The communication initialisation exploits the Bitcraze API and is performed with the drone on the ground.
The SMs share the same set of states:  \emph{ground} (drone on the ground waiting for the takeoff command or after task completion), \emph{hovering} (drone reaching nominal height and waiting the start), \emph{ready} (drone ready for the task to start or already performing it), \emph{stopping} (drone landing). Since the channels are initialised from the OCU, which knows when the connection is established, it can directly command takeoff after the initialisation phase. We configure the OCU as the master that sends commands for synchronisation and triggers state transitions on the drone’s SM. We achieve it by sending messages to/from the drone with a number encoding the current state or the command for state transition. The OCU sends a command to trigger the state transition and waits for a message with the desired state from the drone before updating its own state until state \emph{ready} is reached.
We perform the obstacle detection task when in this latter state. The state messages and detection data are sent over the wireless link every time a new image is received on the Wi-Fi channel and processed via the object detector. The drone uses the link also to send the pose estimate every \qty{10}{ms} for logging purposes and subsequent analysis. The drone communicates information regarding the execution of the navigation task and when it has reached an intermediate or final waypoint. This information is encoded with a number and then matched on the OCU side. Once the final waypoint is reached, the OCU triggers a task stop command and the drone lands.

\subsection{Scene acquisition, transmission, and dataset creation}

The acquisition and streaming task is performed on the AI-deck with code running on its GAP8 processor. The drone SM executes only on the principal drone MCU and the streaming starts as soon as the drone is turned on using an adaptation of an open-source Bitcraze script. Initially, the camera and the Wi-Fi link are set up during an initialisation phase; thereafter, the images are collected, encoded (JPEG), and transmitted during the main loop. The images are collected with the ULP greyscale camera mounted on the AI-deck at full resolution ($320$x$240$ px, $8$ bpp, RAW format), stored in a pre-allocated slot of the L2 memory, and then sent over Wi-Fi using Bitrcraze's \gls{CPX} protocol. In this case, CPX uses the TCP transport protocol with a \gls{MTU} of \qty{1022}{bytes}. The drone acts as a Wi-Fi access point thanks to the u-blox NINA module on the AI-deck. The real-time task requirements determine the number of images acquired per second; therefore, one of our experiments aims to identify all the delays involved in acquisition and streaming.

\begin{figure}[!ht]
  \begin{center}
    \includegraphics[width=\linewidth]{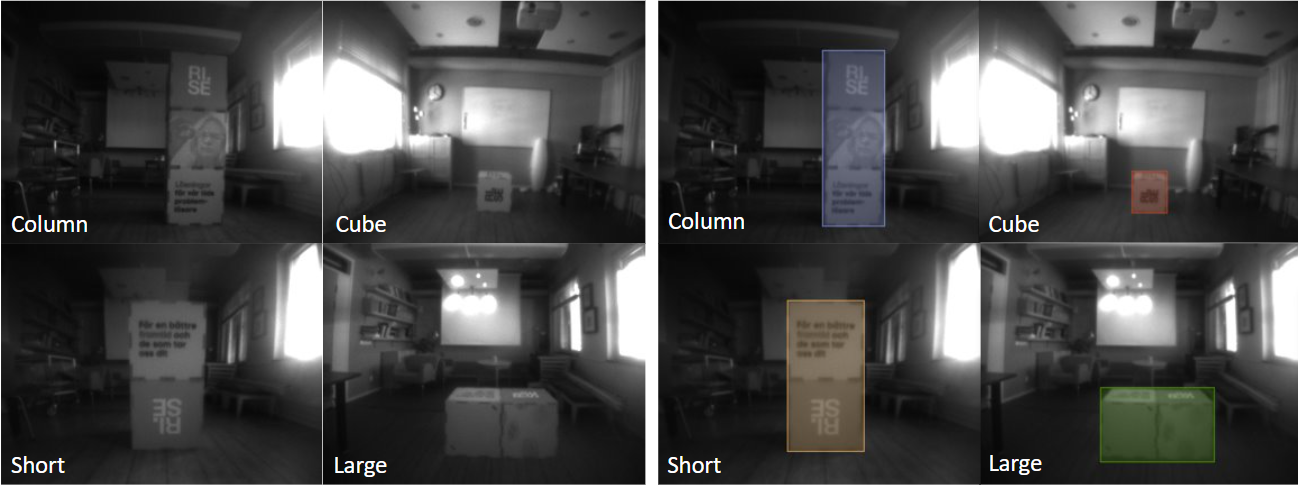}
  \end{center}
  \vspace{-3mm}
  \caption{\textbf{Dataset Example.} The original images (left) are collected from the drone. The same images after labelling (right).}
  \vspace{-4mm}
  \label{fig:dataset_example}
\end{figure}

We consider $4$ different classes of obstacles (see \autoref{fig:dataset_example}). The boxes composing the obstacles are each of the dimension of $50\times50\times50$ cm, and the predefined shapes are \emph{cube}, as a single box, \emph{short}, with two stacked boxes, \emph{large}, with two boxes placed next to each other, and \emph{column}, obtained by stacking three boxes.
We built a balanced dataset collecting and labelling $1000$ RAW images. The images were collected with the drone flying at constant height while moving back, front and sideways to cover the area in front and near the obstacle while having it on the image plane under many points of view from as far as $\sim$\qty{3}{m} from the obstacle to $\sim$\qty{20}{cm}. We collected samples for each object in multiple configurations and with different light conditions. The varying pitch and roll angles during collection caused scene rotation in the images. For these reasons, we did not apply specific augmentation during training. Our data collection was carried out in an office environment with a cluttered background. 
We also included samples with highly occluded obstacles. Given the similarity between the objects, it is difficult, if not impossible, even for a human to guess the correct class when only a portion of the obstacle is visible. Even though this could bring noise into the training process, we wanted the model to detect the presence of an obstacle also in case of occlusions prioritising the detection and safety rather than the correct classification.

\subsection{Deep-learning based Obstacle Detection}
\label{ssec:model}
We adopted a deep learning architecture for object detection called SSD MobileNet V2 renowned for being a fast object detector tailored for mobile and resource-constrained platforms \cite{sandler_mobilenetv2_2018}.  
The original model comes from the Tensorflow Model Zoo, has been trained on COCO 17 dataset and has an input dimension of $320\times320$ px. We performed fine-tuning to allow the detection of our $4$ custom classes of obstacles.  
We randomly selected $80\%$ samples for training, $10\%$ for validation and $10\%$ for testing. We trained the model for $20000$ steps using a batch size of $16$ samples and \emph{random horizontal flip} and \emph{ssd random crop} as data augmentation options. Apart from the learning rate that we set at $0.08$, we used the standard training configuration included in the pre-trained model. Specifically, we used a momentum optimiser with a coefficient of $0.9$ and a cosine decay schedule for the learning rate. Moreover, the total loss is split into two parts, one for classification and one for localisation. The standard ones are the weighted sigmoid focal loss and the weighted smooth L1 loss, respectively. 

The obstacle detection follows the pseudocode in Algorithm \ref{alg:remote_inf}. 
The received image \texttt{img} is first normalised as $\frac{\texttt{img} - \mu}{\sigma}$ where $\mu=\sigma=127.5$. Using the APIs of the Tensorflow \texttt{Interpreter} we get the bounding box (BB) coordinates, class ID and the confidence score for each detected obstacle. Then, we look for the obstacle with the greatest confidence score, above a \textit{threshold} of $0.5$. The final \textit{detection} contains the ID, confidence score, and BB coordinates of the most likely obstacle. The coordinates \textit{xm,ym,xM,yM} identify the top-left and bottom-right corners of the BB.

{
\begin{algorithm}
\small
\caption{Obstacle Detection}\label{alg:remote_inf}
\begin{algorithmic}

\Function{runInference}{\textit{img}}
\State $ img \gets $ \Call {normalize}{$img$}
\State $ boxes,classes,scores \gets $ \Call {predict}{$img$}
\State $max_{score} \gets threshold $
\ForAll{$i$ in $range(len(scores))$} 
    \If{$(scores[i] > max_{score}) \And (scores[i] \leq 1)$}
        \State $ max\_score \gets scores[i] $
        \State $ xm,ym,xM,yM \gets $ \Call {getCoord}{$boxes[i]$}
        \State $ obst_{id} \gets $ \Call{getObstId}{$classes[i]$}
        \State $ detection \gets [obst_{id}, scores[i], xm, ym, xM, yM] $
    \EndIf
\EndFor
\State \Return $detection$
\EndFunction

\end{algorithmic}
\end{algorithm}
}

\subsection{Reactive Planning}
\label{ssec:planning}
In our reactive \emph{planning} algorithm the outcome is a command in longitudinal velocity and yaw rate that is forwarded to the low-level hierarchical PID originally implemented in the Crazyflie firmware. Overall, we compute a repulsive action that keeps the drone away from obstacles in a semi-unknown, or partially known environment. The known components are the dimensions of the area to explore and the shape of the obstacles (not their location). From the area dimensions, a plan is retrieved in terms of waypoints to traverse. We limited our study to the traversal of the space between two waypoints; the method can then be replicated for more target locations.

We use 2D coordinates in the path planning for simplicity. The drone starts from a waypoint $P_A$ to reach a target $P_B$. We denote the estimated pose as $\hat{P}$.

\begin{figure}[!ht]
  \begin{center}
    \includegraphics[width=\linewidth]{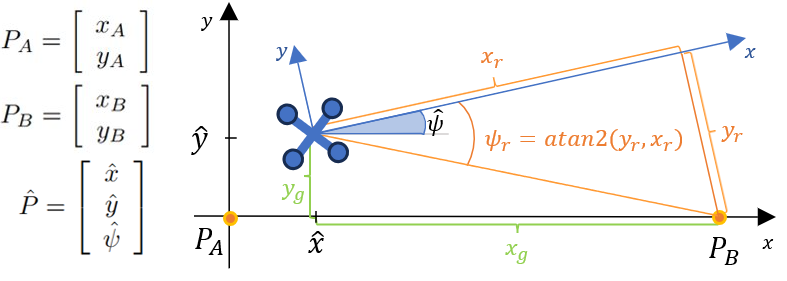}
  \end{center}
  \caption{\textbf{Reference frames and Variables of Interest.} The image illustrates the global (black) and body fixed (blue) reference frames. The variables of interest are the two waypoints $P_A$ and $P_B$, the drone's estimated pose in the global frame $(\hat{x},\hat{y},\hat{\psi})$, the coordinates of the distance from $P_B$ in the global frame $(x_g,y_g)$ and in the body fixed frame $(x_r,y_r)$, and the heading offset $\psi_r$.}
  \vspace{-3mm}
  \label{fig:ref_frames}
\end{figure}

The origin of the global reference frame is in $P_A$ and with the x-axis directed towards $P_B$. We also consider a relative reference frame body fixed to the drone and with the x-axis pointing in the UAV forward direction. In each iteration, we compute the relative distance from the drone to the target point and obtain the angle offset $\psi_r$ from the current yaw to the heading that would lead the drone directly to the target. \autoref{fig:ref_frames} illustrates this setting.

\begin{figure}[!ht]
  \begin{center}
    \includegraphics[width=\linewidth]{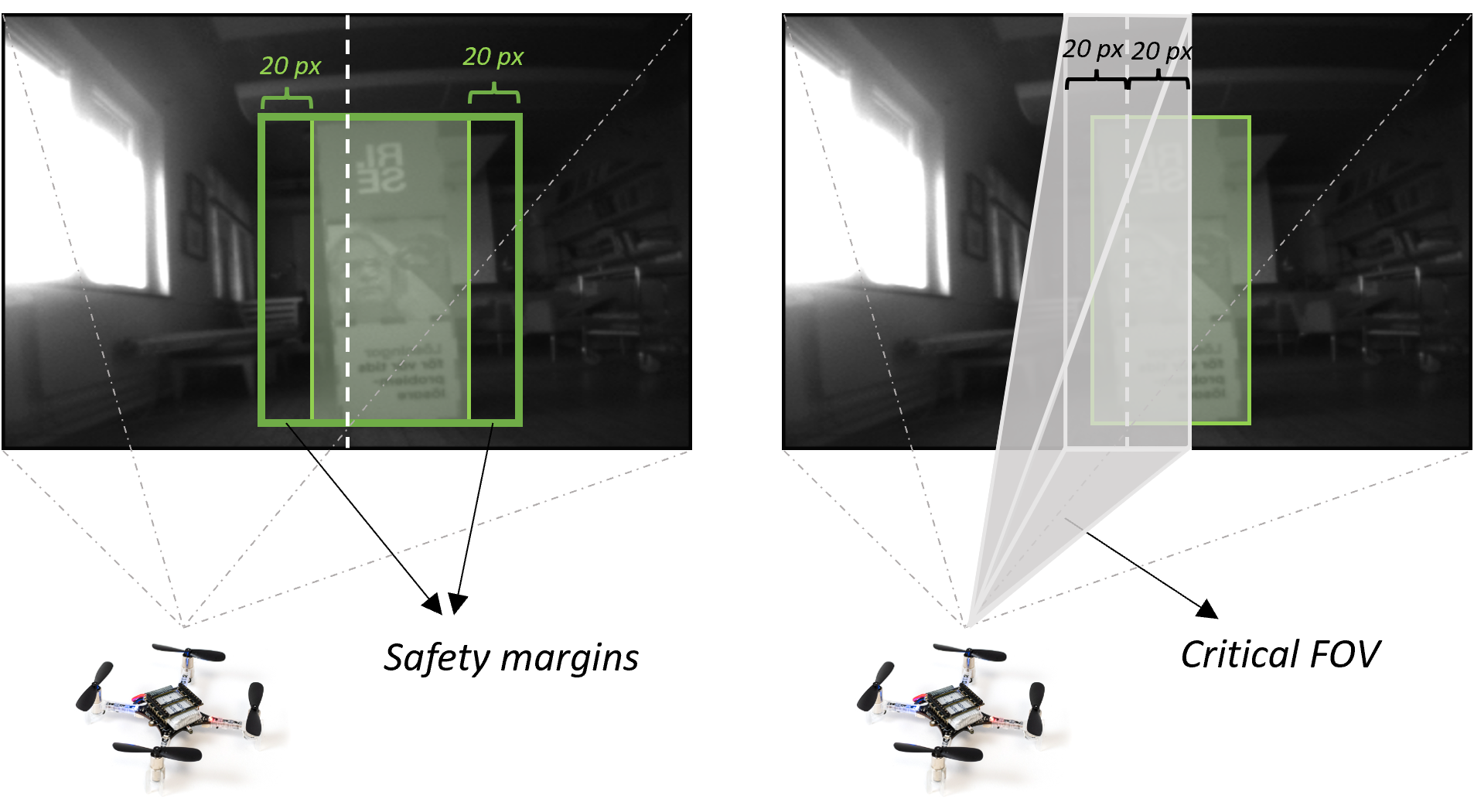}
  \end{center}
  \caption{\textbf{Safety margin and critical FOV.} (left) Inflation of the detected BB by a safety margin of \qty{20}{px} on each side. (right) Concept of critical FOV, i.e. the FOV enclosing a window of \qty{40}{px} centered in the middle of the image plane.}
  \label{fig:critical_fov}
  \vspace{-3mm}
\end{figure}

\begin{figure}
    \centering
    \subfloat[][]{\includegraphics[width=0.425\columnwidth]{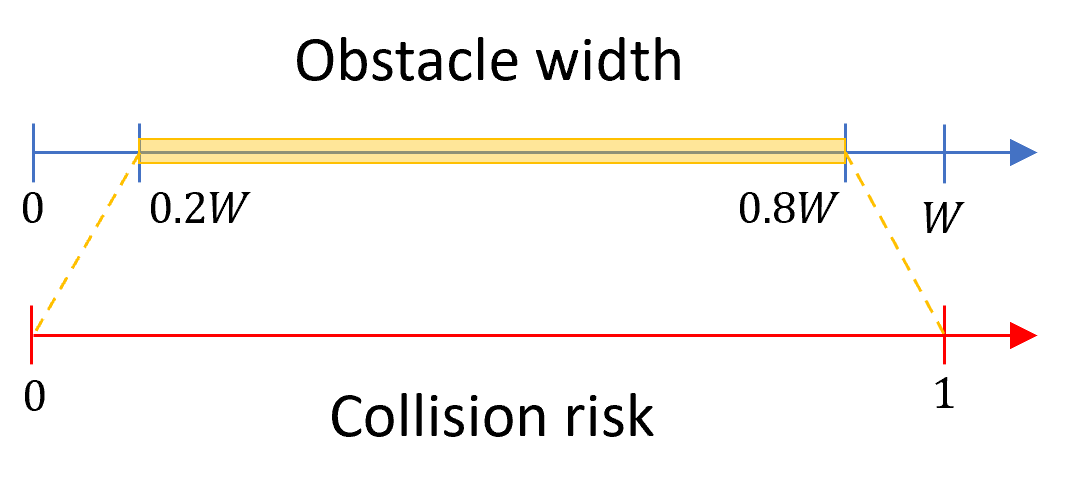} \label{fig:coll_risk_mapping}}
    \hspace{-2mm}
    \subfloat[][]{\includegraphics[width=0.525\columnwidth]{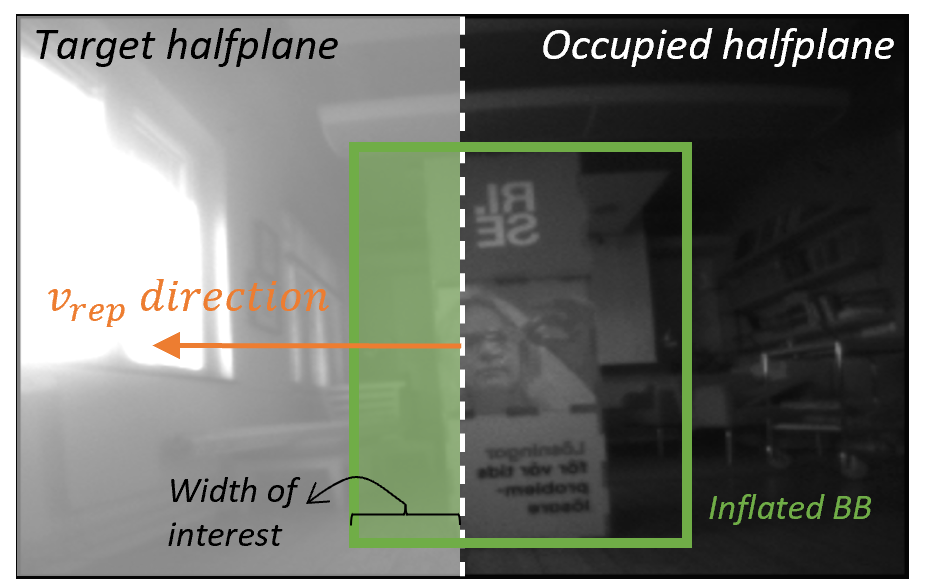} \label{fig:rep_velocity}}
      \vspace{-2mm}
    \caption{(a) \textbf{Rescaling of the Collision Risk.} The Collision risk is derived from the width of the BB. $W$ represents the width of the image plane. (b) \textbf{Repulsive Velocity.} The repulsive velocity $v_{rep}$ is directed towards the less-occupied halfplane by the inflated BB. Its module is proportional to the width of interest, i.e. the width of the portion of the BB lying in the target halfplane.}
    \vspace{-3mm}
\end{figure}

The drone receives the BB information and computes a \emph{collision risk} based on the width of the obstacle as follows:
\begin{itemize}[itemsep=0mm,leftmargin=2mm]
    \item We add a safety margin of \qty{20}{px} to the obstacle edges in the horizontal direction. This is to enhance safety and take into account the drone dimensions. There is a risk of collision only if the inflated BB intersects the vertical line at the centre of the image plane. In terms of detecting a possible collision, this is equal in concept to the drone looking for obstacles in a window of width 40px and centred in the middle of the image plane. We call critical FOV the one enclosing the window (\autoref{fig:critical_fov}).
    \item We normalise and remap the collision risk in the range $[0,1]$ when the width of the detected obstacle is $[20\%,80\%]$ of the total image width \textbf{\emph{W}} as shown in \autoref{fig:coll_risk_mapping}. The lower limit is to prevent the drone from reacting too early to far objects, which are smaller in the image plane. The upper limit further enhances safety. 
\end{itemize}

We use the collision risk to modulate the longitudinal velocity. However, since it is monotonically increasing with the width of the obstacle, we transform it to a monotonically decreasing variable called \emph{safety factor} $S_t$. This factor indicates the level of safety during flight at each iteration of the algorithm. 
\[ S_t = (coll_{risk,t} - 1)^2 . \]
Since the safety factor is derived from the BB and, thus, depends on the deep learning model accuracy, we apply exponential smoothing to its value to filter out the model noise. The smoothing coefficient $\alpha=0.5$ is empirically determined.
\[ S_t = \alpha S_t + (1-\alpha)S_{t-1}, \quad \alpha=0.5  \]
The smoothed safety factor multiplies the longitudinal velocity. The desired velocity command $v_d$ is constrained to a maximum speed $v_{max}$ of \qty{1}{m/s}.
\[ v_{d,t} = min( v_{max}, \sqrt{x_r^2+y_r^2} \times S_t \times \bigg|\frac{\psi_{r,t}}{180}-1\bigg| )   .\]
Where $x_r$, $y_r$ and $\psi_r$ are as in \autoref{fig:ref_frames}. The last multiplier is to reduce the forward velocity when $\psi_r$ is high, i.e., the drone is heading far away from the target.

The repulsive action is computed only if an obstacle is present and intersects the critical FOV. The first step in the computation of the desired yaw rate $\dot{\psi}_d$ is the calculation of a repulsive action that allows the drone to change heading and avoid obstacles, as shown in \autoref{fig:rep_velocity}. We call it \emph{repulsive velocity} $v_{rep}$ and derive it from the width of the inflated BB. In particular:
\begin{itemize}[itemsep=0mm,leftmargin=2mm]
    \item It is directed towards the halfplane that has more free space, thus avoiding the region of the image plane that is occupied the most by the obstacle.
    \item Its value corresponds to the width of the part of the obstacle lying in the selected halfplane. The action is proportional to the “quantity” of the obstacle to avoid. We normalise $v_{rep}$ in a range $[0,1]$ corresponding to a width from $[0,W/2]$ where $W/2$ is the width of the halfplane.
\end{itemize}

 The width of interest, $WoI$, is shown in \autoref{fig:rep_velocity}. The repulsive velocity is given as:
\[ v_{rep,t} = k_{vel} \times \frac{WoI_t}{W/2} . \]
The coefficient $k_{vel}$ determines the strength of the repulsive action and directly impacts the obstacle avoidance manoeuvre. Its influence on the navigation task was studied in the experimental phase (see \autoref{sec:results}).
Since $v_{rep}$ is derived from the BB, similarly to the case of the safety factor, we smooth its value with a low pass filter and a smoothing coefficient $\beta=0.5$.
\[ v_{rep,t} = \beta v_{rep,t} + (1-\beta)v_{rep,t-1}, \quad \beta=0.5 .\]
From the repulsive velocity we compute the \emph{repulsive yaw} $\psi_{rep}$ as:
\[  \psi_{rep,t} = atan2(v_{rep,t}, v_{d,t}) . \]
Finally, we obtain the desired yaw rate command $\dot{\psi}_d$ by considering a balance between the repulsive yaw $\psi_{rep,t}$ and the attractive action, identified as the relative yaw $\psi_{r,t}$.
\[ \dot{\psi}_{d,t} = (\psi_{r,t}S_t + \psi_{rep,t})/dt, \quad dt=0.2 . \]
The multiplication with the safety factor limits the contribution of the attractive action when the risk of collision is high. The term $dt$ is a time variable needed to transform the desired yaw (in the parenthesis) in an angular velocity. We chose $dt=0.2s$ to approximate the period corresponding to the frequency at which a new command is sent to the controller.
The desired yaw rate is constrained to a maximum angular velocity $\dot{\psi}_{max} = 60 deg/s$. Algorithm \ref{alg:planning_alg} presents the pseudocode of the planning algorithm that has a $\mathcal{O}(1)$  complexity.
{
\begin{algorithm}
\small
\caption{Reactive Planning}\label{alg:planning_alg}
\begin{algorithmic}

\Procedure{planningTask}{$drone\_pose,obstacle$}
\State $P \gets drone\_pose$
\State $T \gets current\_target$
\State $O \gets obstacle$

\If{\Call{targetReached}{$P,T$}}
    \State \Call{stayOnTarget}{$T$}
    \State \Call{updateTarget}{$T$}
    \State \Return
\EndIf

\If{$ O \neq null $}
    \State $coll_{risk,t} \gets $ \Call{collisionRisk}{$O$}
    \State $S_t \gets (coll_{risk,t}-1)^2$
    \If{$ coll_{risk,t} > 0$}
        \State $ v_{rep,t} \gets k_{vel} \times \dfrac{WoI_t}{W/2} $
    \EndIf
\EndIf
\State $ v_{rep,t} \gets \beta v_{rep,t} + (1-\beta)v_{rep,t-1} $
\State $ S_t \gets \alpha S_t + (1-\alpha)S_{t-1} $

\State $ x_r,y_r \gets $ \Call {globalDistance}{$P,T$}
\State $ dx_{rel},dy_{rel} \gets $ \Call{relativeDistance}{$P,x_r,y_r$}
\State $ \psi_{r,t} \gets atan2(dy_{rel},dx_{rel}) $

\State $ v_{d,t} \gets min( v_{max}, \sqrt{x_r^2+y_r^2} \times S_t \times \big|\frac{\psi_{r,t}}{180}-1\big| ) $

\State $ \psi_{rep,t} \gets atan2(v_{rep,t}, v_{d,t}) $

\State $ \dot{\psi}_{d,t} \gets (\psi_{r,t}S_t + \psi_{rep,t})/dt $
\EndProcedure

\end{algorithmic}
\end{algorithm}
  \vspace{-3mm}
}

\section{Performance Evaluation and Results}
\label{sec:results} 

\subsection{Delay and Timing}
The latency between perception and actuation is the main component affecting the reactive capabilities of the system, hence the successful and safe navigation of the quadrotor. To evaluate such latency, we studied the streaming end-to-end delay, streaming rate, inference time, and planning time.

\begin{figure}[!ht]
  \begin{center}
    \includegraphics[width=\linewidth]{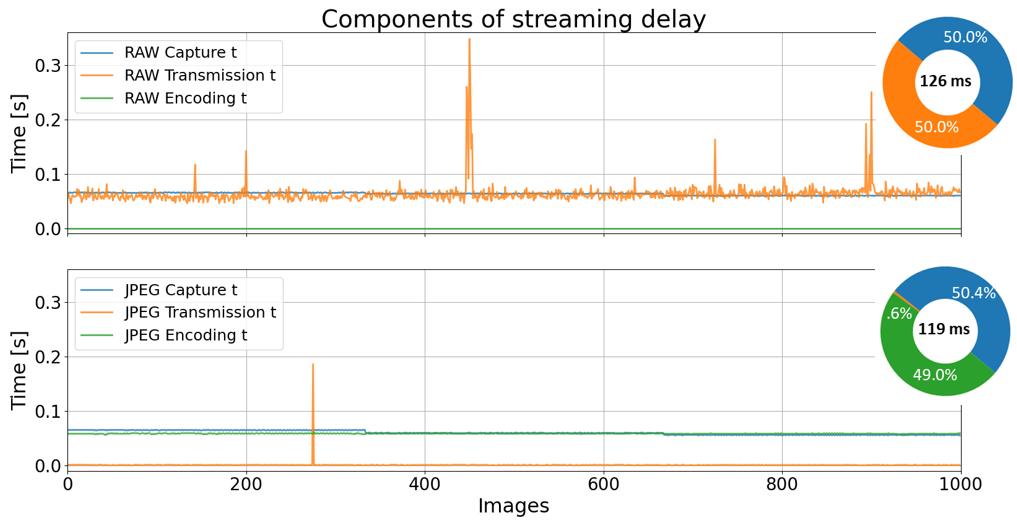}
  \end{center}
    \vspace{-2mm}
  \caption{\textbf{Components of streaming delay for RAW and JPEG format.} Trace of the capturing, encoding and transmission times for $1000$ streamed RAW and JPEG images. The pie charts show the impact of each component and the average total delay between consecutive transmissions.}
    \vspace{-2mm}
  \label{fig:streaming_components}
\end{figure}

We assessed the end-to-end delay comprising of the image capture (raw format), encoding (JPEG), and transmission components as shown in \autoref{fig:streaming_components}. 
The transmission duration depends on the amount of data to send and, therefore, on the image format. Indeed, an image weighs $\sim$ \qty{79}{kB} in RAW format or $\sim$ \qty{6}{kB} if encoded in JPEG format, requiring at least $79$ or $6$ CPX packages respectively. The lower transmission time for JPEG frames is countered by the encoding time. The peak observed in the transmission time for JPEG format is attributed to the changes in channel condition and associated loss of packets causing retransmission. In both cases, the overall average time is comparable and around \qty{120}{ms}.

Since the above does not account for the propagation and processing time, we also characterised the end-to-end delay using on-screen timer method~\cite{Singhal_VTC}. Averaging the results from $25$ measurements for each image format, we got $T_{stream}^{RAW} = (\mu : 326, \sigma : 48)\,\text{ms}$ and $T_{stream}^{JPEG} = (\mu : 223, \sigma : 54)\,\text{ms}$. The greater amount of data to transmit with RAW images impacts the propagation and processing (reading) time, accounting for $\sim 100 \,\text{ms}$ more in the end-to-end streaming delay. 

\begin{figure}[!ht]
  \begin{center}
    \includegraphics[width=\linewidth]{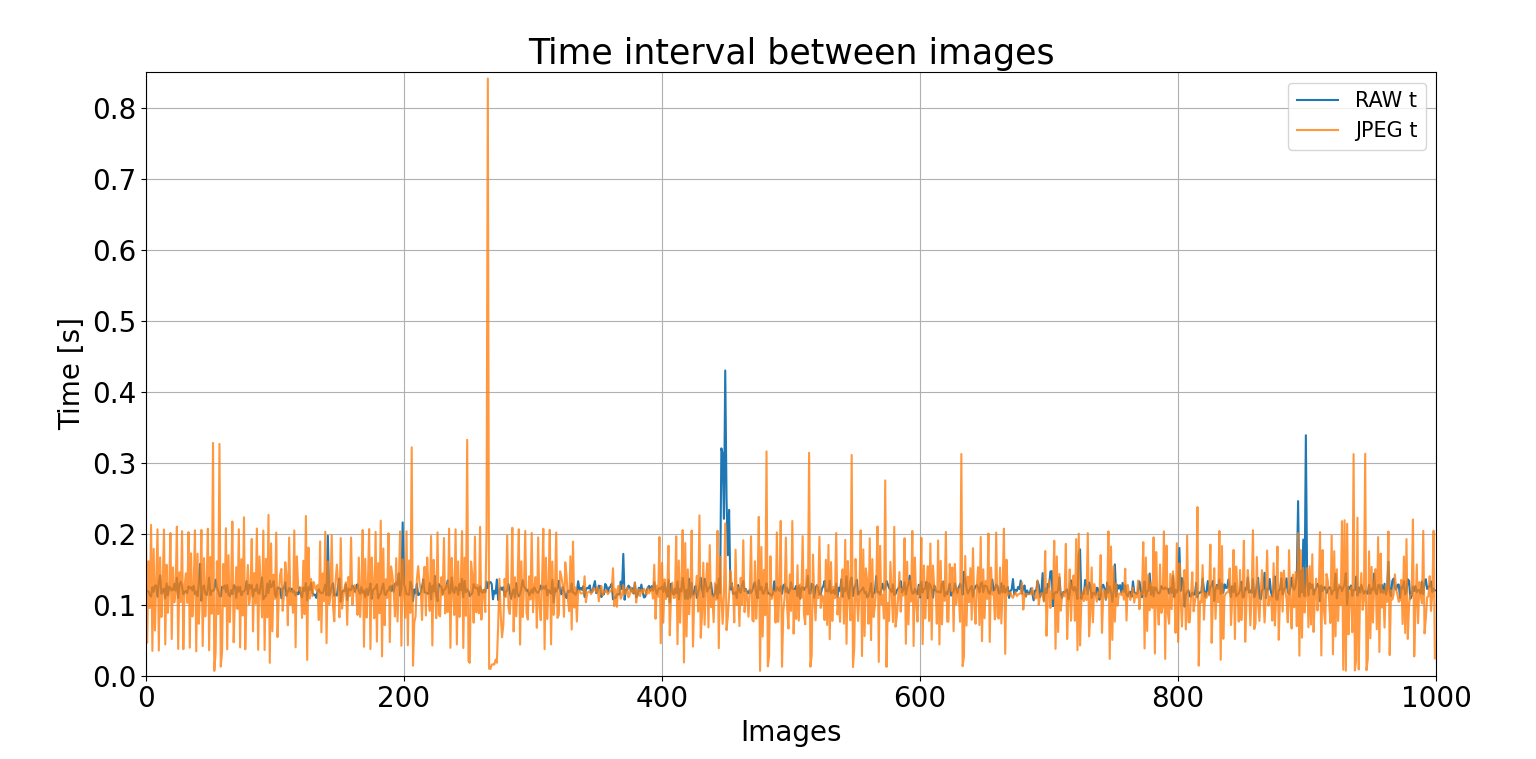}
  \end{center}
    \vspace{-2mm}
  \caption{\textbf{Streaming Rate or Arrival Time.} The figure shows the traces of the amount of time elapsed between the reception of two following images. The samples refer to $1000$ images collected in both formats.}
    \vspace{-2mm}
  \label{fig:arrival_time}
\end{figure}

\autoref{fig:arrival_time} shows the arrival time of consecutive images and we observe a higher variability for the JPEG compared to the RAW format. This is due to the processing of the images on the OCU side, likely for decoding purposes. We observe lower intervals following the peaks in the JPEG arrival time. This indicates that when the processing time is higher than the transmission process, the laptop still receives the following images and buffers them. When one frame is processed, another is already available in the buffer to be read and stored in memory. Comparing \autoref{fig:arrival_time} and \autoref{fig:streaming_components} we notice that the variability in the arrival time for the RAW images is directly connected to the variability in the transmission time.
Considering the average time of arrival, we obtain \qty{123}{ms} and \qty{119}{ms} for RAW and JPEG. This results in $\sim$\qty{8}{FPS} on average for both formats.

We have also evaluated the inference time, i.e., the time the SSD MobileNet V2 model takes to process the input and return an output with the detection information. The model was running on the CPU of a ThinkPad-T450s laptop. The obtained average inference time over $1000$ measurements was \qty{51}{ms}. Therefore, on average, an output is ready before the arrival of the following image, thus not decreasing the rate at which the output is sent to the drone for the planning task. Additionally, we have also measured the planning time, which refers to the time the drone takes to compute a command when an input message with object detection information is received. This quantity is on average lower than \qty{1}{ms}.

\subsection{Object Detection}
A reliable perception of the environment is fundamental to avoiding obstacles. We based our assessment of the object detection model performance on the COCO \gls{mAP} metric~\cite{wang_parallel_2022} and a custom metric derived from it. We first evaluated the model on the test split of the original dataset. Then, we tested it on images streamed in real-time from the drone during the experiments for obstacle avoidance. This second time we assessed model performance with images in both RAW and JPEG format. 
As already mentioned, in the proposed framework we wanted to prioritise safe navigation to correct classification, indeed the planning algorithm works independently from the class of the detected BB. Therefore, we studied the model's performance by grouping all the obstacles into a single generic class.
The overall COCO mAP averaged over all IoU thresholds for the generic obstacle class is summarised in \autoref{tab:map_overall}.

\begin{table}[!ht]
  \begin{center}
  {\small
    \caption{mAP results for the generic obstacle class}
      \vspace{-2mm}
    \label{tab:map_overall}
    \begin{tabularx}{\linewidth}{|X|*{3}{>{\centering\arraybackslash} m{0.5in}|}}
      \hline
       & Test Set & RAW & JPEG \\
      \hline
      \textbf{Overall mAP $@.5:.95$} & $85.72$ & $60.80$ & $62.56$ \\
      \hline
    \end{tabularx}
  }
  \end{center}
    \vspace{-2mm}
\end{table}

\begin{figure}[!ht]
  \begin{center}
    \includegraphics[width=0.9\linewidth]{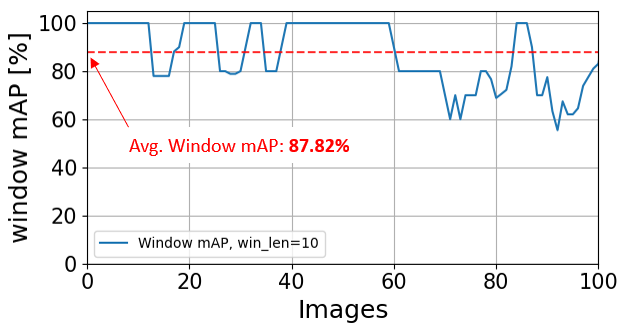}
  \end{center}
   \vspace{-3mm}
  \caption{\textbf{Window mAP metric} on a 10-frames window during real obstacle avoidance experiments.}
  \vspace{-3mm}
  \label{fig:window_map}
\end{figure}

We note the following about the inference performance:
\begin{itemize}[itemsep=0mm,leftmargin=2mm]
    \item The model achieved good localisation and most detections would allow safe navigation given that the bounding boxes cover the correct width of the visible obstacle.
    \item The mAP score drops if evaluated at high IoU thresholds ($\geq0.9$). However, an IoU of $0.5$ is often considered good and is for example the only value used in the Pascal VOC challenge.
    \item The model rarely produced erroneous detections on objects being part of the cluttered background (only few FPs detected). Also, we detected a significant amount of TPs, accounting for $84\%$ of all positive samples.
\end{itemize}
Most importantly, we have formulated the planner to be robust to noise in the detection by introducing a low-pass filter on the safety factor and reactive action, which reduces the impact of false predictions and enhance safe navigation. In other words, if the model correctly detects an obstacle in most frames but fails in others, it can keep memory of the danger in terms of a safety factor and still avoid collisions. 
The existing metrics in literature do not consider this memory aspect. Therefore, we have formulated a new custom metric, called window mAP, obtained by computing the moving average of the mAP at IoU$=0.5$ in a window of $10$ consecutive frames. The result of the \emph{window mAP} evaluation is presented in \autoref{fig:window_map}.

We note that the score remains above $70\%$ mostly, and the average score over all the windows results in a window mAP of $87.82\%$.

\subsection{Exploration and Obstacle Avoidance}

\begin{figure}
    \centering
      \subfloat[\label{fig:complete_setting}]{
      \includegraphics[width=0.43\columnwidth]{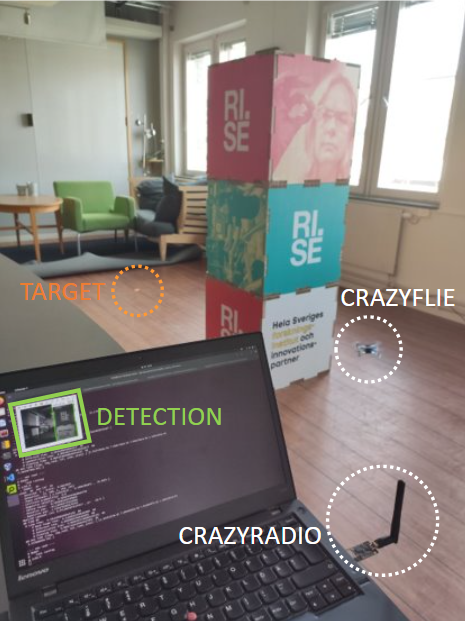}  
    }\hspace{-2mm}
      \subfloat[\label{fig:exp_setting_large}]{
      \includegraphics[width=0.51\columnwidth]{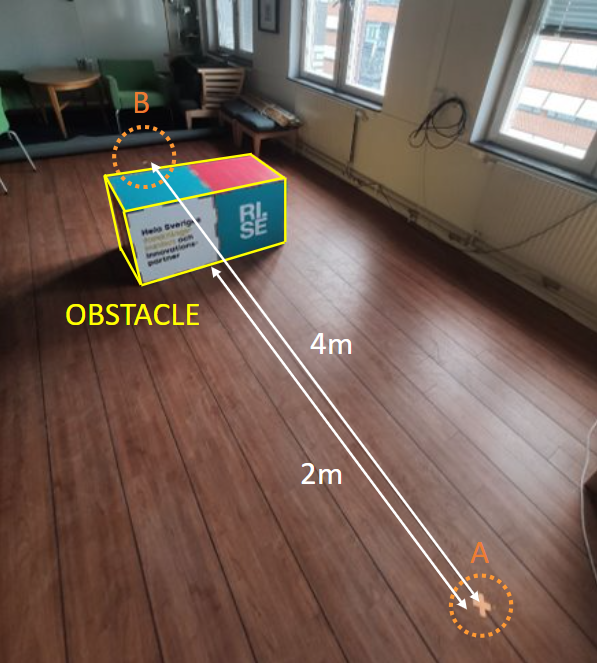}
      }
    \caption{(a) \textbf{Experimental setting with a radio} Drone navigates towards the target point while avoiding the obstacle using the detections sent via the radio channel. (b) \textbf{Experiment Setting.} Position of start point, destination and obstacle during the experiments.}
    \label{fig:exp_setting_all}
    \vspace{-2mm}
\end{figure}

We further evaluated the effectiveness of the developed nano-drone framework to complete the navigation task safely by scaled-down experiments in a semi-unknown environment. We evaluated the reactive planning method by means of repetitive tests with different objects while analysing the effect of the parameter $k_{vel}$ ruling the strength of the repulsive action.

The drone had to traverse the space between two waypoints, \qty{4}{m} apart, going from position A to position B. An obstacle was placed between the waypoints, at \qty{2}{m} from the start and blocking the direct way to destination, as illustrated in Fig.~\ref{fig:exp_setting_all}. The experiment was conducted only for the obstacles of class \emph{short} and \emph{large} and repeated $5$ times to evaluate the  reliability. We identify the best setting of parameters for the reactive planning algorithm among $k_{vel}=[0.5,0.7,1,1.5,2]$ comparing the effect of the different values by computing task completion time, length of the path and number of collisions.

\begin{figure}[!b]
    \begin{center}
      \includegraphics[width=\linewidth]{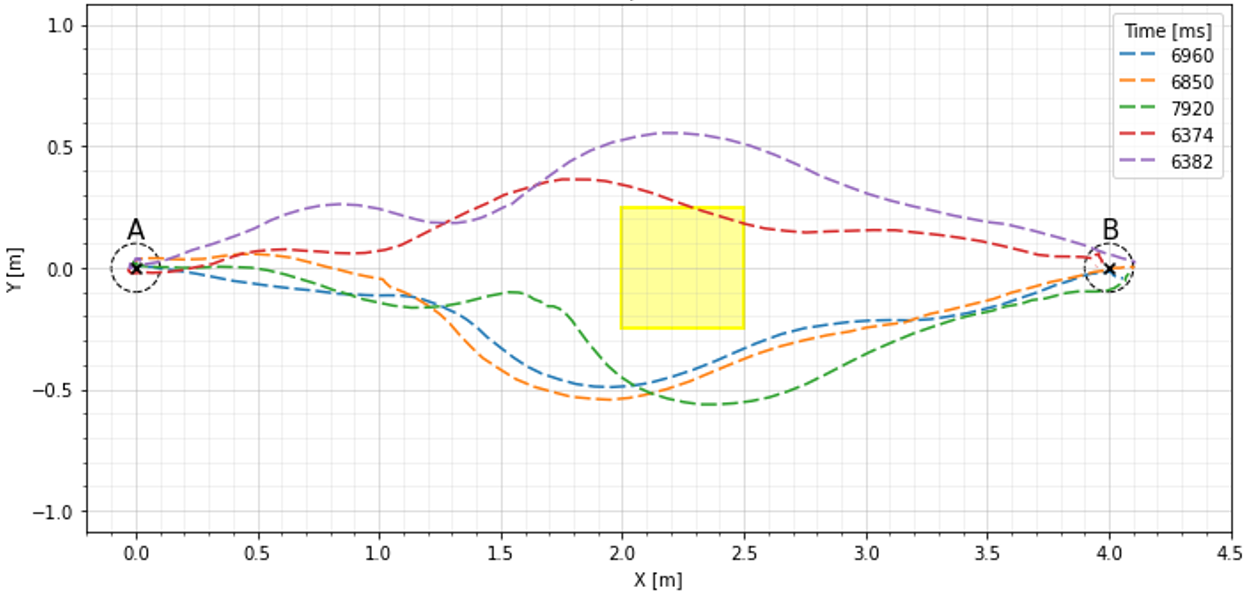}
    \end{center}
      \vspace{-2mm}
    \caption{\textbf{Obstacle Avoidance best experiments - class \emph{short}.} Drone paths with class \emph{short} at $k_{vel}=1.5$.}
      \vspace{-2mm}
    \label{fig:short_avoidance}
\end{figure}

\begin{figure}[!b]
    \begin{center}
     \includegraphics[width=\linewidth]{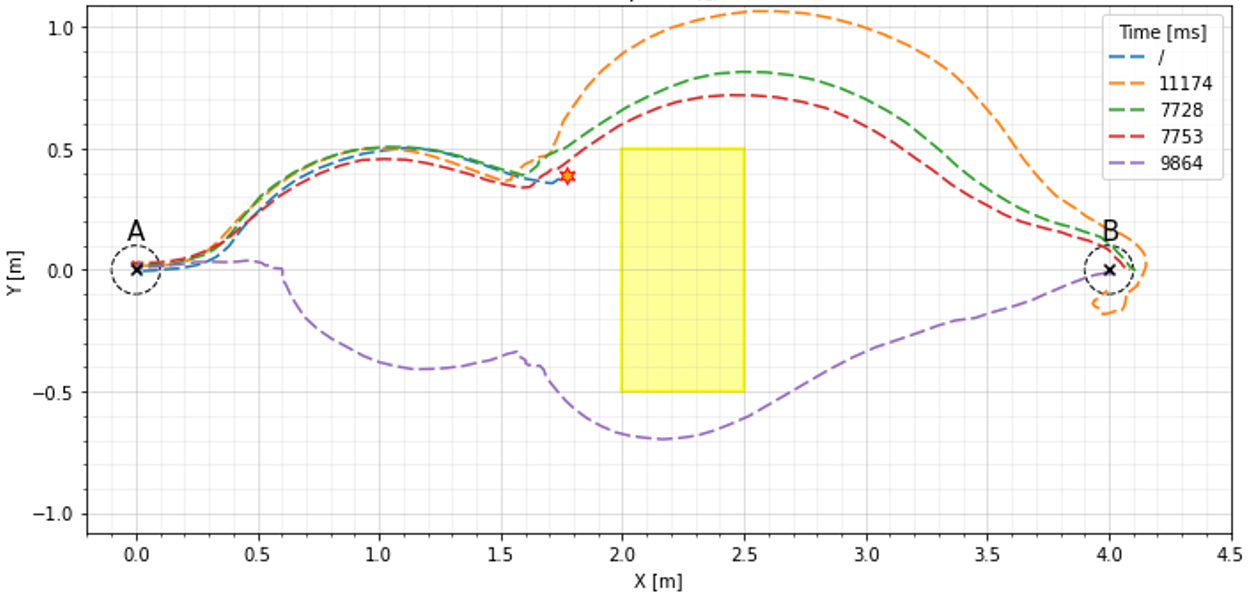}
    \end{center}
      \vspace{-3mm}
    \caption{\textbf{Obstacle Avoidance best experiments - class \emph{large}.} Drone paths with class \emph{large} at $k_{vel}=0.5$.}
      \vspace{-3mm}
    \label{fig:large_avoidance}
\end{figure}

\begin{figure}[!ht]
  \begin{center}
    \includegraphics[width=0.7\linewidth]{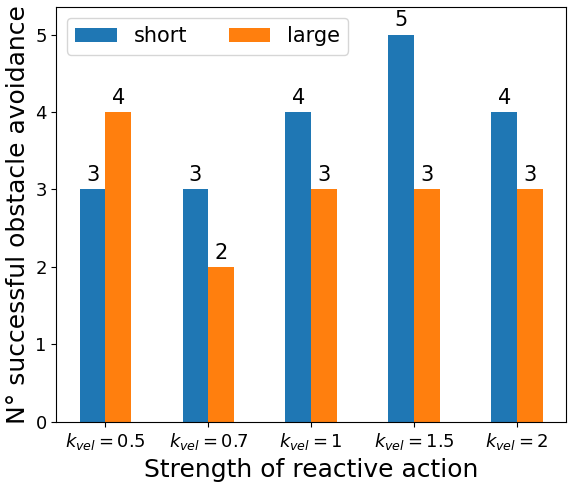}
  \end{center}
    \vspace{-2mm}
  \caption{\textbf{Successful Obstacle Avoidance Manoeuvres.}}
    \vspace{-2mm}
  \label{fig:successful_oa}
\end{figure}

In the following, \autoref{fig:short_avoidance} and \autoref{fig:large_avoidance} illustrate a top view of the resulting paths from the best batch of experiments for the obstacles of type \emph{short} and \emph{large}, obtained with $k_{vel}=1.5$ and $k_{vel}=0.5$, respectively. 
In all completed tasks, the drone reached the dashed circle (radius \qty{10}{cm}) and landed inside it. We note one unsuccessful test highlighted with a star at the end of the path. In some runs, the displayed path overlaps the obstacle (in yellow) but there is no collision reported. This is because of a misalignment between the onboard state estimation and the known global position of the obstacle. The lack of ground truth data for the drone's trajectories prevented analysing the minimum distance to the cubes during the avoidance manoeuvre.
During the experiments some tests failed due to causes such as losing a propeller while flying, hitting other objects (not obstacles under test) in the environment or other unknown malfunctions. \autoref{fig:successful_oa} reports a bar graph of all the successful obstacle avoidance manoeuvres for each parameter’s value and each obstacle. In the graph, a task is considered successful if the drone managed to go over the obstacle, no matter of a later malfunction.

In general, with the \emph{short} obstacle, we obtained a higher number of avoided collisions accounting for the $76\%$ of all runs compared to the $64\%$ achieved for class \emph{large}. This statistic shows that the wider object is more challenging to avoid, which is understandable given the nature of the reactive approach and the fact that the obstacle is $10$ times larger than the drone size. The proposed percentages consider the experiments with all the tested values of $k_{vel}$, but our goal was to identify the best settings. We further present the data regarding the average time and average travelled distance in every batch of tests (only for the completed runs), in \autoref{tab:comp_time} and \autoref{tab:path_length}. We note that in terms of the completion time, the best ($k_{vel}=1.5$) and second best values apply for both classes

\begin{figure}[h]
  \subfloat[][Avg completion time (ms) \label{tab:comp_time}]{ 
    \begin{tabular}{l|l|l}
        $k_{vel}$ & \emph{short} & \emph{large} \\
        \hline \hline
        $0.5$ & $7138$ & $9129$ \\
        $0.7$ & $8050$ & $8495$ \\
        $1$ & $7163$ & $8934$ \\
        $1.5$ & $\mathbf{6897}$ & $\mathbf{6939}$ \\
        $2$ & $\mathit{7063}$ & $\mathit{6983}$
    \end{tabular} }
  \hfill
  \subfloat[][Avg path length (m) \label{tab:path_length}]{
    \centering
    \begin{tabular}{l|l|l}
        $k_{vel}$ & \emph{short} & \emph{large} \\
        \hline \hline
        $0.5$ & $\mathbf{4.34}$ & $4.83$ \\
        $0.7$ & $4.67$ & $\mathbf{4.53}$ \\
        $1$ & $4.42$ & $\mathit{4.68}$ \\
        $1.5$ & $\mathit{4.37}$ & $4.69$ \\
        $2$ & $4.40$ & $4.73$
    \end{tabular} 
  }
  \caption{\textbf{Average completion time and path length.} The table shows the average completion time (in milliseconds) and path length (in metres) for the successful runs. The best value is in bold and the second best value in italic.}
    \vspace{-4mm}
  \label{tab:time_and_length}  
\end{figure}

Overall, we can consider $k_{vel}=1.5$ the best parameter setting under all perspectives. Indeed, averaging the results of the two classes of obstacles for all the parameters, it obtained the higher percentage of successful avoidance manoeuvres ($80\%$), the lower completion time (\qty{6918}{ms}) and shorter path length (\qty{4.53}{m}). However, the separate results for the two obstacles indicate potential advantages of a solution with settings that adapt dynamically to the obstacle class. 

\section{Conclusions and Future work}
\label{sec:conclusionsAndFutureWork}
We have developed a safe navigation ISCC solution for a Crazyflie nano-drone based on AI-aided visual reactive planning for obstacle avoidance in a  partially known environment. 
The modules of the safe navigation pipeline include the deep learning based obstacle detection model and the path planning algorithm. We experimentally evaluated the latency involved in the communication between the drone and the OCU, the performance of the object detector, and the parameter based path planning algorithm. We studied the impact of each designed component on the safe navigation task. We successfully achieved a processing rate of \qty{8} {FPS} and an object detection performance of 60.8 of COCO mAP. Through the field experiments we verified the solution feasibility and achieved the safe navigation of the drone flying at a top speed of \qty{1}{m/s} while steering away from an obstacle placed in an unknown position and reaching the target destination. Our framework successfully implements the real-time navigation task requirements for the resource constrained nano-drone platform. 
In future work, we will explore multi-point navigation, trajectory optimisation, and the tradeoff between an edge-offloaded obstacle detection and AI model execution aboard the drone.

\section*{Acknowledgments}
\label{sec:ack}
\small
This work was made possible by the industrial internship at RI.SE, Kista, Sweden and the double degree program Autonomous Systems and Intelligent Robots, coordinated by EIT Digital. The program has received funding from the European Community’s DIGITAL Programme under Grant Agreement No. 101123118 (SPECTRO).

\bibliography{main}

\end{document}